\documentclass[lettersize,journal]{IEEEtran}
\usepackage{amsmath,amsfonts}
\usepackage{algorithmic}
\usepackage{algorithm}
\usepackage{array}

\ifCLASSOPTIONcompsoc
    \usepackage[caption=false, font=normalsize, labelfont=sf, textfont=sf]{subfig}
\else
\usepackage[caption=false, font=footnotesize]{subfig}
\fi

\usepackage{textcomp}
\usepackage{stfloats}
\usepackage{url}
\usepackage{verbatim}
\usepackage{graphicx}
\usepackage{cite}

\usepackage{bbding} 
\usepackage{multirow}
\usepackage{threeparttable} 

\begin{document}
\title{Tac2Structure: Object Surface Reconstruction \\ Only through Multi Times Touch}
\author{Junyuan Lu, Zeyu Wan, and Yu~Zhang{*} 
\thanks{This work was supported by the National Natural Science Foundation of China (Grant No.62088101), the National Key Research and Development Program of China under Grant 2021ZD0201403, the Project of State Key Laboratory of Industrial Control Technology, Zhejiang University, China (No.ICT2021A10), the Open Research Project of the State Key Laboratory of Industrial Control Technology, Zhejiang University, China (No.ICT2022B04). \textit{(Corresponding author: Yu Zhang.)}}
\thanks{J. Lu, Z. Wan and Y. Zhang are with the State Key Laboratory of Industrial Control Technology, College of Control Science and Engineering, Zhejiang University, Hangzhou 310007, China (e-mail: junyl@zju.edu.cn; zeyuwan@zju.edu.cn; zhangyu80@zju.edu.cn)}
}

\IEEEpubid{
    \begin{minipage}{\textwidth}
    \ \\[30pt]
    \copyright 2023 IEEE\\
    Personal use of this material is permitted.  Permission from IEEE must be obtained for all other uses, in any current or future media, including reprinting/republishing this material for advertising or promotional purposes, creating new collective works, for resale or redistribution to servers or lists, or reuse of any copyrighted component of this work in other works.
    \end{minipage}
} 


\maketitle
\begin{abstract}
Inspired by humans’ ability to perceive the surface texture of unfamiliar objects without relying on vision, the sense of touch can play a crucial role in robots exploring the environment, particularly in scenes where vision is difficult to apply, or occlusion is inevitable. 
Existing tactile surface reconstruction methods rely on external sensors or have strong prior assumptions, making the operation complex and limiting their application scenarios. 
This paper presents a framework for low-drift surface reconstruction through multiple tactile measurements, Tac2Structure.
Compared with existing algorithms, the proposed method uses only a new vision-based tactile sensor without relying on external devices.
Aiming at the difficulty that reconstruction accuracy is easily affected by the pressure at contact, we propose a correction algorithm to adapt it.
The proposed method also reduces the accumulative errors that occur easily during global object surface reconstruction.
Multi-frame tactile measurements can accurately reconstruct object surfaces by jointly using the point cloud registration algorithm, loop-closure detection algorithm based on deep learning, and pose graph optimization algorithm. 
Experiments verify that Tac2Structure can achieve millimeter-level accuracy in reconstructing the surface of objects, providing accurate tactile information for the robot to perceive the surrounding environment.

\end{abstract}


\section{Introduction}
\IEEEPARstart{T}{he} ability of robots to complete manipulation tasks strongly depends on accurate information about the contact object. 
Visual information is generally sufficient to perceive, track, or recognize objects in most scenes\cite{newcombe2011kinectfusion}. 
However, visual methods have many disadvantages, such as low accuracy and occlusion. Commonly used visual depth cameras usually have centimeter-level accuracy for depth measurements. 
In robot manipulation tasks, occlusion is inevitable when the robot is physically in contact with objects, causing visual methods to fail. 
To solve this problem, a novel image-based tactile sensor, such as Gelsight\cite{yuan2017gelsight} or GelSlim\cite{donlon2018gelslim} is added to the end of the manipulator to sense the structural properties of the contact object, which can enhance the robot’s knowledge of the scene. This sensor finely senses the texture of the contact surface and outputs high-resolution images, which can be converted into 3D point clouds using a photometric stereo algorithm\cite{woodham1979photometric}.
Inspired by the human reading Braille, which combines various pieces of information through multiple touches, robots can fuse multi-frame point clouds and reconstruct the surface of objects using the point cloud registration algorithm.

\IEEEpubidadjcol

However, most current studies obtained the pose transformation of the sensor through external devices\cite{li2018end,bauza2019tactile} to improve the fusion accuracy, significantly limiting the application scenarios. 
Owing to the influence of sampling pressure\cite{wang2021gelsight}, the reconstructed texture obtained by feature matching\cite{li2014localization} lacks consistency in depth, which produces uneven fracture surface phenomena, leading to significantly reduced reconstruction accuracy. 
In addition, during the reconstruction of large-scale object surfaces whose scale is at least ten times the effective sampling area of the sensor, the accumulative error generated by multi-frame fusion is often unacceptable, seriously affecting the accuracy.

\begin{figure}[t]
\centering
\includegraphics[width=\linewidth]{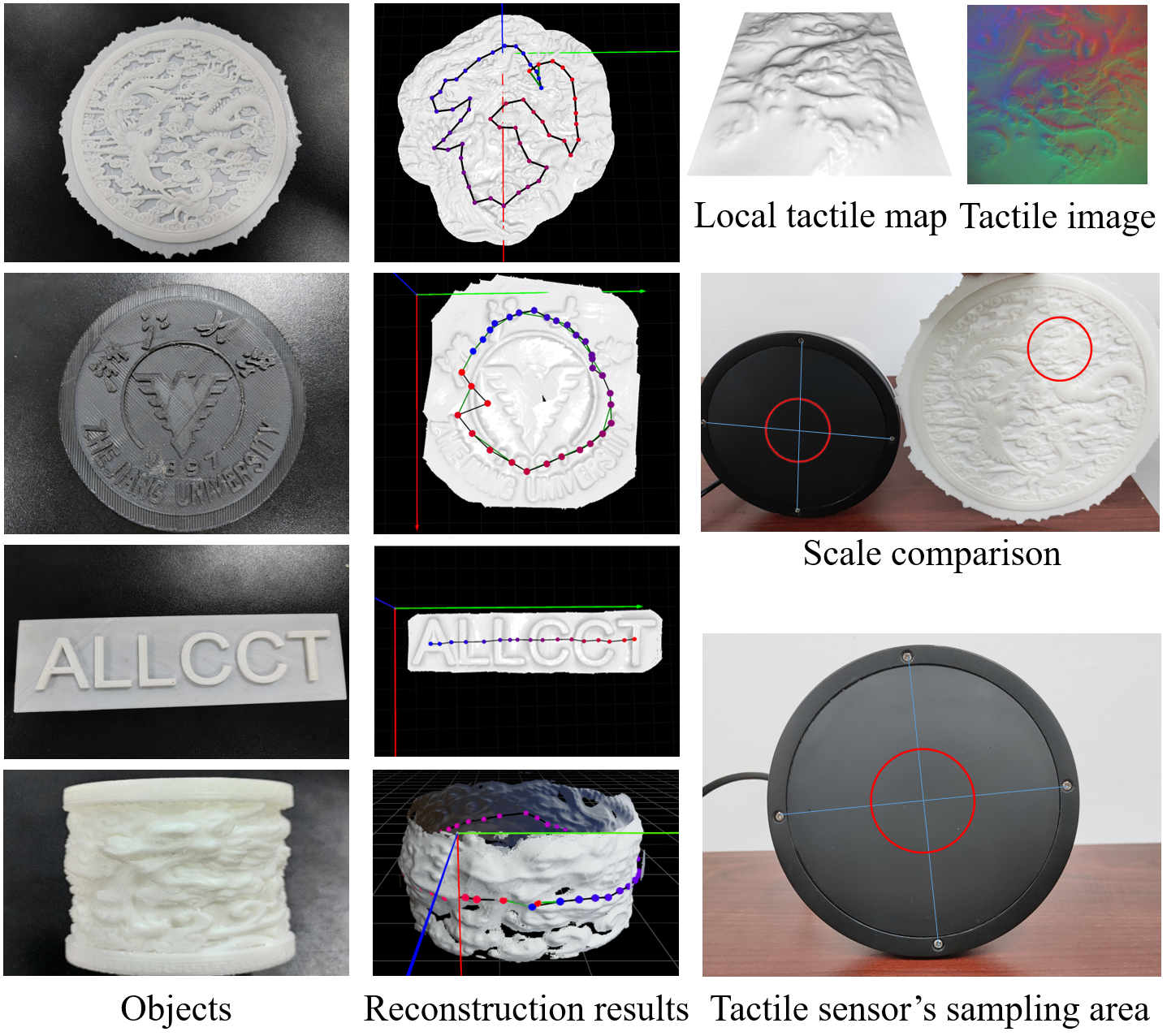}
\caption{\textbf{Tac2Structure algorithm.} 
This work solves the problem of accurately reconstructing the surface of objects. Without relying on any external device, we can reconstruct the surface of an unfamiliar object with low drift through multiple touch measurements.}
\label{fig:ourwork}
\end{figure}

To solve these problems, we propose a novel framework called Tac2Structure for simultaneous sensor localization and object surface reconstruction based on a tactile sensor only. 
First, we use the point cloud registration algorithm to estimate the pose transformation of the sensor between multi-sampled frames such that the global reconstruction process does not rely on external devices. 
Second, an adaptive pressure correction algorithm is proposed to enhance the consistency of the depth between frames and reduce the difficulty of the actual sampling operation. 
Finally, we utilize the loop-closure detection method based on deep learning and the pose graph optimization algorithm to reduce the accumulative error and improve accuracy.

To the best of our knowledge, this work is the first to achieve low-drift reconstruction of object surfaces without relying on external observation equipment. The proposed algorithm can provide accurate surface texture information of the contact objects for the robot to make subsequent decisions, enhancing the cognitive ability of the robot for unknown objects and improving the robustness of the robot in performing tasks in unfamiliar scenes or scenes with visual occlusion.
In general, the proposed algorithm makes the following contributions:
\begin{itemize}
\item{Egomotion estimation: can accurately estimate the sensor’s egomotion without relying on the accurate observations provided by other external sensors.}
\item{Pressure adaptive: can overcome the sensor’s inherent disadvantage of accuracy affected by the sampling pressure.}
\item{Low-drift global reconstruction: can reduce the accumulative error and achieve millimeter-level accuracy.}
\end{itemize}

\begin{figure*}[!t]
\centering
\includegraphics[width=\linewidth]{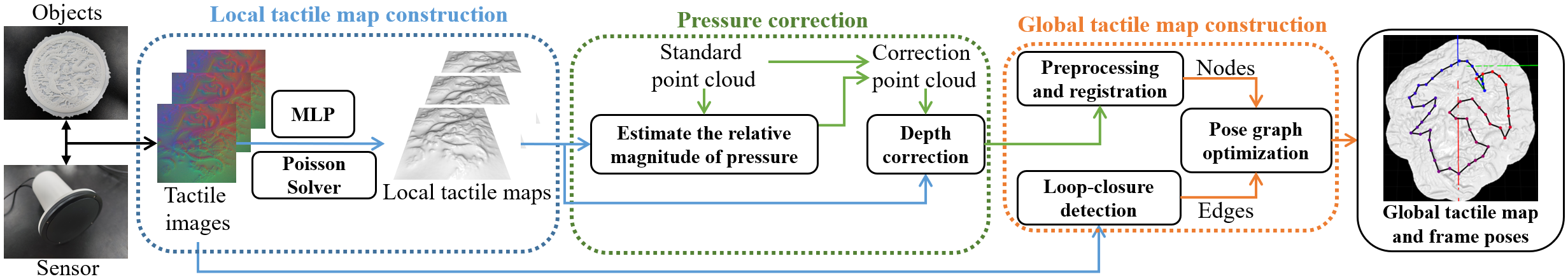}
\caption{\textbf{Tac2Structure framework.} 
To reconstruct the surface of an object, we design a system comprising three modules: 1) local tactile map construction, 2) pressure correction, and 3) global tactile map construction. Finally, we can get the surface reconstruction result and each sampling frame’s sensor pose. \textbf{As shown in the global tactile map, some legend settings will be used: 1) the order of sampling: starting from the blue point and ending at the red point; 2) the black and green lines: odometry and loop-closure edges of pose graph, respectively.}}
\label{fig1-framework}
\end{figure*}

\section{Related Work}
Owing to the lack of tactile perception, traditional robots can only achieve rough effects on tasks, such as interacting with humans, grasping, and manipulation. Furthermore, unacceptable operational errors may occur in complex and unfamiliar scenes or scenes with visual occlusions. 
To solve these problems, researchers have begun to add tactile sensors to the ends of robots to perceive objects better and improve their operational abilities\cite{wettels2009multi,dong2017improved,huh2020dynamically,suresh2021tactile}.  
Image-based tactile sensors are gradually becoming mainstream tactile sensors owing to their advantages of high resolution and low cost.

Here, we use a tactile sensor similar to Gelsight\cite{yuan2017gelsight}, which uses a soft gel and built-in camera to collect high-resolution tactile signals.  
The image represents the raw data type of the output of the sensor. Because of the structure of the three LED sources separated inside the sensor, the image can be converted into a 3D point cloud using a photometric stereo algorithm\cite{woodham1979photometric}.

Recently, several researchers have successfully utilized this sensor to assist robots in task performance. 
Ma et al. used the inverse FEM algorithm to estimate the force distribution on a contact surface\cite{ma2019dense}. 
Chaudhury et al.\cite{chaudhury2022using} and Dikhale et al.\cite{dikhale2022visuotactile} jointly used vision and tactile information to estimate the pose between the robot and object.
The above research only employs the information of a single frame of tactile data, which implies that the potential of tactile sensors has still not been fully exploited. Our work combines multi-frame sensor data to construct a more accurate object surface texture.
The studies by \cite{li2014localization},\cite{li2018end} and \cite{bauza2019tactile} are related to this study.

Li et al.\cite{li2014localization} first proposed the concept of a tactile map regarding the multi-frame tactile map registration problem as an image mosaicing problem. They registered multi-frame local tactile maps using feature-based matching techniques, assuming that tactile maps have the same scale and that there will be no out-of-plane rotation. 
To construct the global tactile map, they coarsely averaged the depth values of the overlapping areas between the frames.
However, when pressed manually to sample, making the forces consistent requires a high sampling operation. Inconsistent forces reduce the accuracy of subsequent global tactile map construction. 
Because the fusion process only considers the $SE$(2) pose transformation, it cannot be applied to surfaces with undulations, such as cylindrical surfaces. 
Because the incremental reconstruction process easily produces accumulated errors, this simple method is unsuitable for large-scale object surface reconstruction. 
Our method automatically adapts to the pressure without limiting the operation of the sampling process. 
During global tactile map construction, instead of being constrained to solve the pose transformation in 2D, our method directly solves it in 3D. 
Our method uses loop-closure detection and pose-graph optimization techniques for large-scale objects to mitigate the accumulative error, significantly reducing the non-negligible accumulated error during the reconstruction process.

In subsequent studies, Li et al.\cite{li2018end} and Bauza et al.\cite{bauza2019tactile} extended the tactile map concept to 3D scenes to reconstruct a 3D object’s surface. 
However, their methods strongly depend on the pose transformation of the tactile sensor provided by external devices, such as sensors on a robot arm with robot kinematics or a motion capture device.
External equipment limits the application scenarios of the algorithm and makes the overall reconstruction system more expensive. 
Our method uses a point cloud registration algorithm to estimate the tactile sensor egomotion between frames, which requires no additional equipment.

Our work is similar to the famous simultaneous localization and mapping (SLAM) framework\cite{zhang2014loam,mur2015orb,mur2017orb,dai2020rgb,DBLP:journals/corr/abs-2111-07723}.
Some of these algorithms use LiDAR information\cite{zhang2014loam}, and some use visual information\cite{mur2015orb,mur2017orb}.
Although tactile images also belong to the visual information category, using only these images to reconstruct surface textures under the pure monocular Visual SLAM framework is challenging. This is because they can only build up-to-scale structures of the sensor pose and environment. 
Our method estimates the accurate scale of the surface texture from raw tactile image data based on the photometric stereo algorithm\cite{woodham1979photometric}.

\section{Methods}
The proposed low-drift surface reconstruction framework is shown in Fig. \ref{fig1-framework}. By receiving multi-frame inputs from a tactile sensor, our framework can accurately estimate the egomotion of the sensor and reconstruct object surfaces with low drift.
Our framework is composed of three key submodels.

\begin{itemize}
\item{\textbf{Local tactile map construction.} 
In this module, given a tactile image, we output its corresponding local tactile map representing the shape of the local contact area. We use the multi-layer perceptron (MLP) model and 2D Poisson solver. We provide a practical toolbox to improve the efficiency of training dataset annotation. Section \ref{section four} explains the detailed process.}
\item{\textbf{Adaptive pressure correction.} 
We mitigate the pressure inconsistencies introduced by manual pressing based on statistical methods. The analysis of the depth error caused by the inconsistent pressure and adaptive correction algorithm are explained in Section \ref{section five}.}
\item{\textbf{Global tactile map construction.}  
Given a series of corrected local tactile maps, we use the point cloud registration algorithm, loop-closure detection algorithm based on deep learning, and pose graph optimization technology to reconstruct the global tactile map. Section \ref{section six} elaborates on the details of this content.}
\end{itemize}

\section{Local tactile map construction\label{section four}}
Similar to the work of Wang et al. \cite{wang2021gelsight}, 
our method uses a multi-layer perceptron and Fast Poisson solver\cite{doernerpythonfast} to estimate the local 3D surface texture from the raw tactile image data. 
Given the focus and length of the paper, we refer readers to their work\cite{wang2021gelsight} for details on the network structure and data collection method. 

\begin{figure}[t]
\centering
    \subfloat[Imprint of pressed ball\label{pressball}]{
    \includegraphics[width=.37\linewidth]{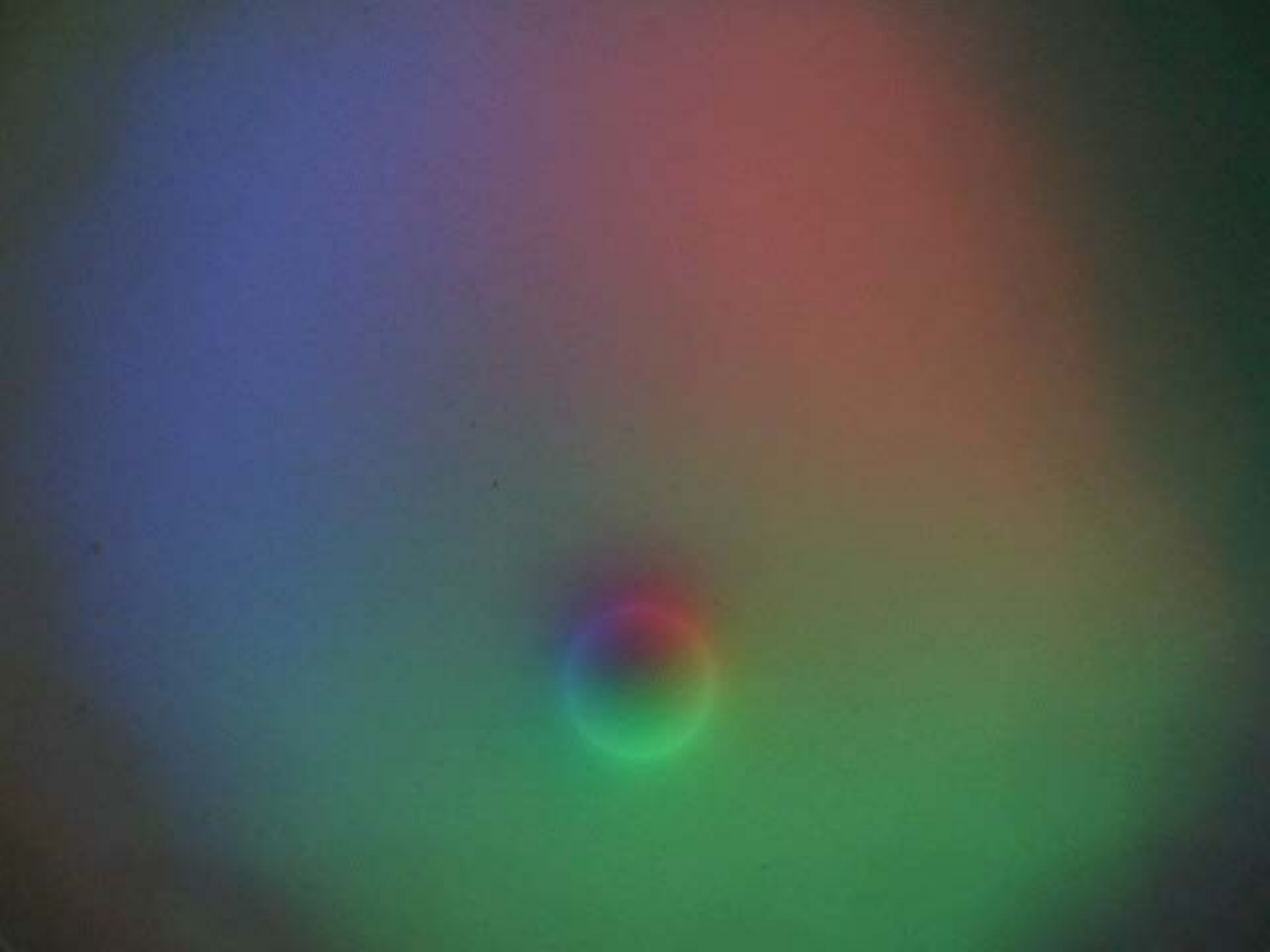}
    }
    \subfloat[Toolbox interface\label{Toolbox_interface}]{
    \includegraphics[width=.48\linewidth]{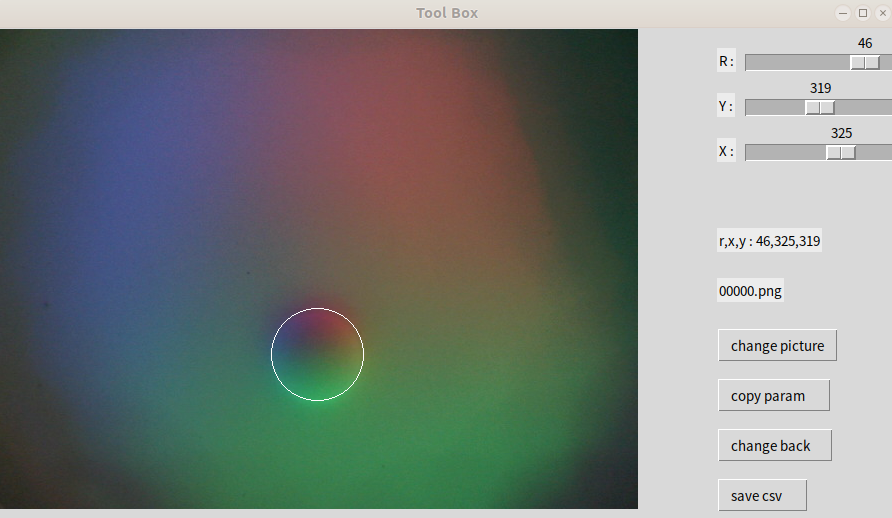}
    }
\caption{\textbf{Training data labeling process.} Different sliders of the toolbox can control the position and radius of the auxiliary circle of the contact circle between the ball and sensor. The position and radius information of the contact circle are obtained when the auxiliary one fits the actual one.}
\label{fig:toolbox}
\end{figure}

After collecting datasets, the difficulty lies in labeling them, which needs to extract the center’s position and the circular area’s radius, as shown in Fig. \ref{fig:toolbox}\subref{pressball}. 
Most existing studies \cite{wang2021gelsight,dong2017improved,romero2020soft} extracted them using the Hough Circle Transform. However, in practice, the detection parameters must be adjusted repeatedly to adapt to different images, which is labor-intensive. 

We implement a toolbox suitable for extracting the required circle parameters to label datasets more efficiently, as shown in Fig. \ref{fig:toolbox}\subref{Toolbox_interface}. Our work is open source, and codes are available at \url{https://github.com/ljy-zju/Tac2Structure.git}.

\begin{figure}[t]
\centering
    \captionsetup[subfloat]{labelformat=empty}
    \includegraphics[width=.9\linewidth]{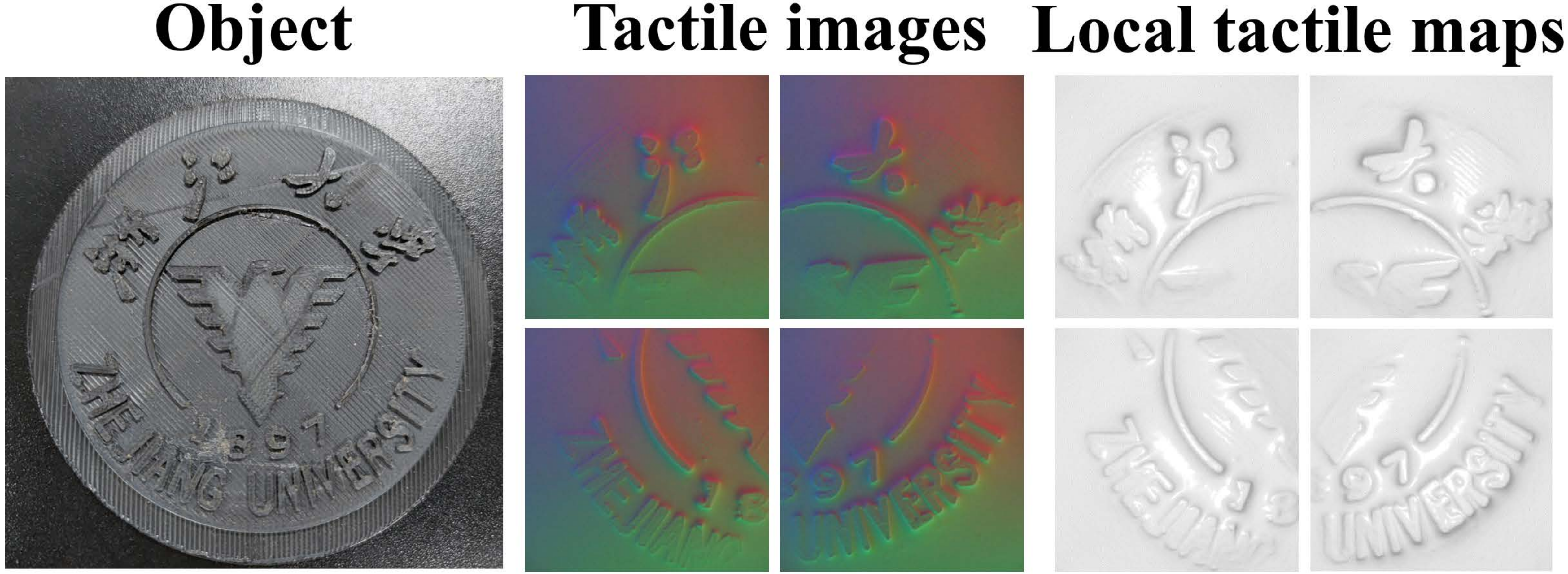}
\caption{\textbf{Local tactile map construction results.} From left to right: visual image, tactile images, and local tactile maps of 3D printed `ZJU’.}
\label{fig:localmapresult}
\end{figure}

The experimental section \ref{subsection:exp_Local} introduces the quantitative results of the local tactile map construction in detail. As shown in Fig. \ref{fig:localmapresult}, we visualize some local tactile maps of the 3D printed `ZJU’ here.

\begin{figure}[b]
\centering
    \subfloat[\label{sensorside}]{
    \includegraphics[width=.3\linewidth]{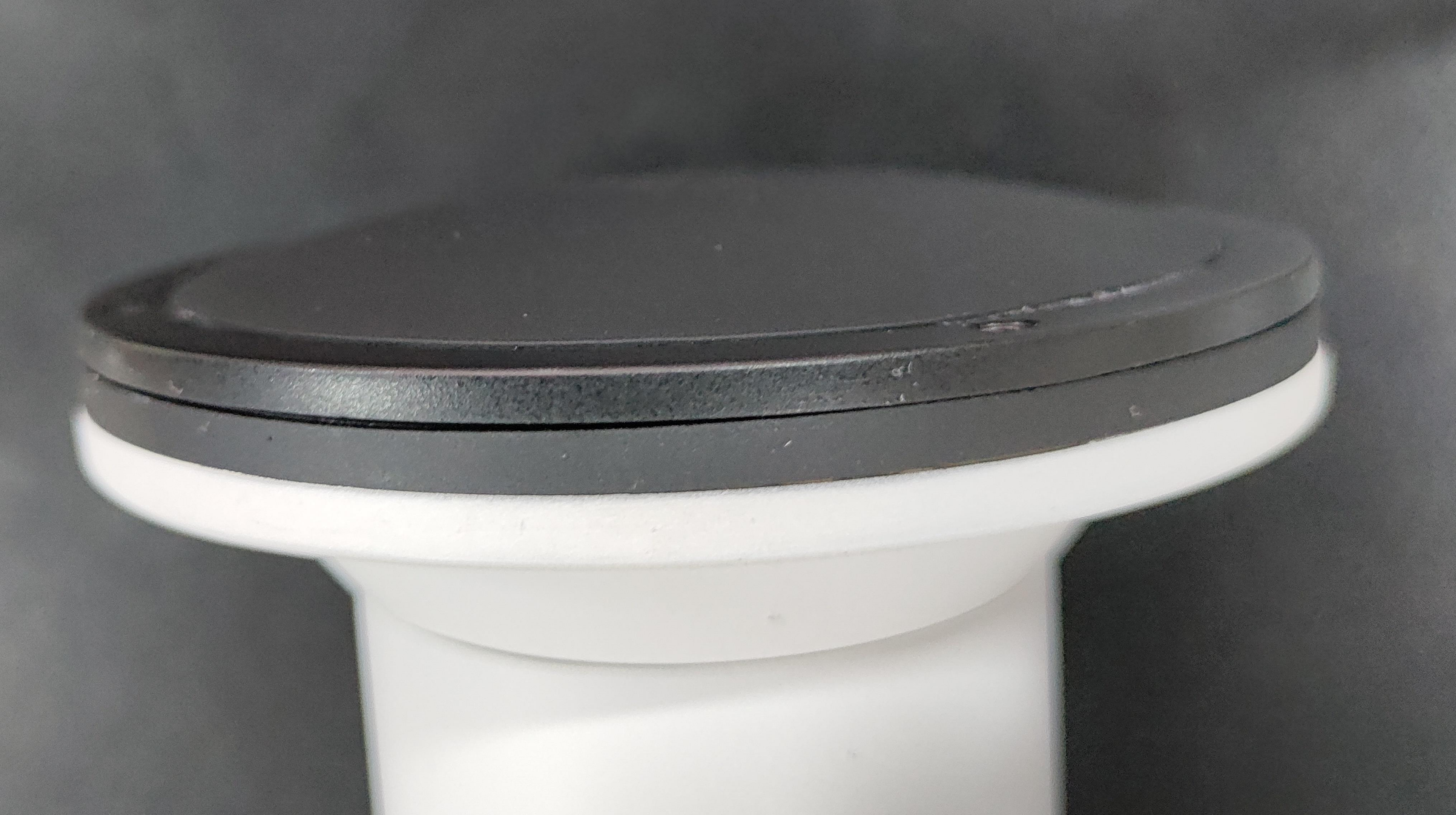}
    }
    \subfloat[\label{pressonplane1}]{
    \includegraphics[width=.33\linewidth]{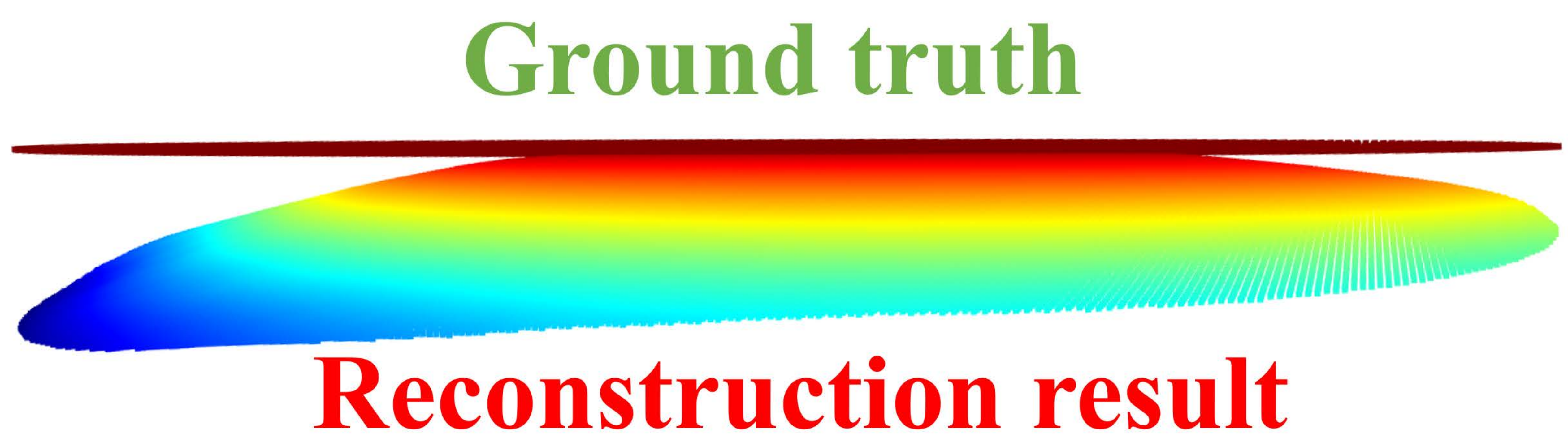}
    }
    \subfloat[\label{pressonplane2}]{
    \includegraphics[width=.33\linewidth]{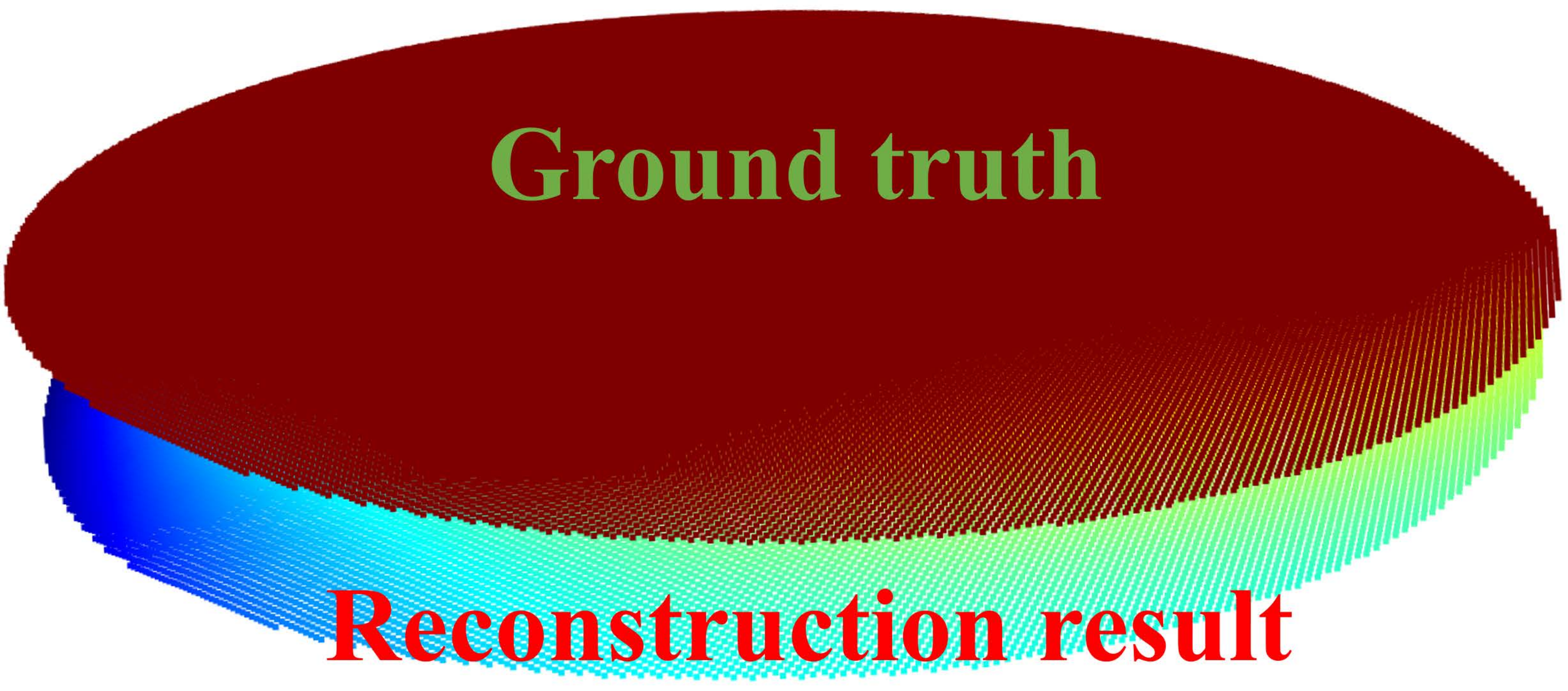}
    }
    \caption{\textbf{Analysis of gel structure of sensor surface.} (a) is the side view of the sensor, which shows a certain spherical arc. (b) and (c) are the side and top views of the local tactile map made by pressing the sensor on a flat desk, respectively. The wine-colored part is the theoretical flat reconstruction result, while the colored part is the actual spherical arc. }
\label{fig:sensor}
\end{figure}

\section{Adaptive pressure correction algorithm\label{section five}}
\subsection{Reconstruction error sources analysis}
\subsubsection{Arc-shaped sensor surface}
The tactile sensor used here is shown in Fig. \ref{fig:sensor}\subref{sensorside}. 
The gel surface is a slight arc, which causes a deviation between the local tactile map construction result and the ground truth, as shown in Fig. \ref{fig:sensor}\subref{pressonplane1}\subref{pressonplane2}. 
The result of the local reconstruction is essentially gel deformation, which is larger in the central region than in the other regions.
Therefore, the local tactile map is characterized by a deep center and shallow periphery, reducing the accuracy of the reconstructed depth.

\subsubsection{Sampling Pressure inconsistency\label{section five-B}}
\begin{table}[t]
    \centering
    \caption{\label{tab:force-and-radius}Different pressures and corresponding fitted spherical radii}
    \resizebox{\linewidth}{!}{
    \begin{tabular}{c|cccccc}
        \hline
        Group & 1 & 2 & 3 & 4 & 5 & 6 \\ 
        \hline
        Pressure  &  \multicolumn{6}{c}{low $\longrightarrow$ high} \\ 
        \hline
        Radius (mm) & 193.3 & 183.1 & 174.9 & 162.2 & 143.7 & 117.8\\
        \hline
    \end{tabular}
    }
\end{table}

\begin{figure}[t]
\centering
    \includegraphics[width=\linewidth]{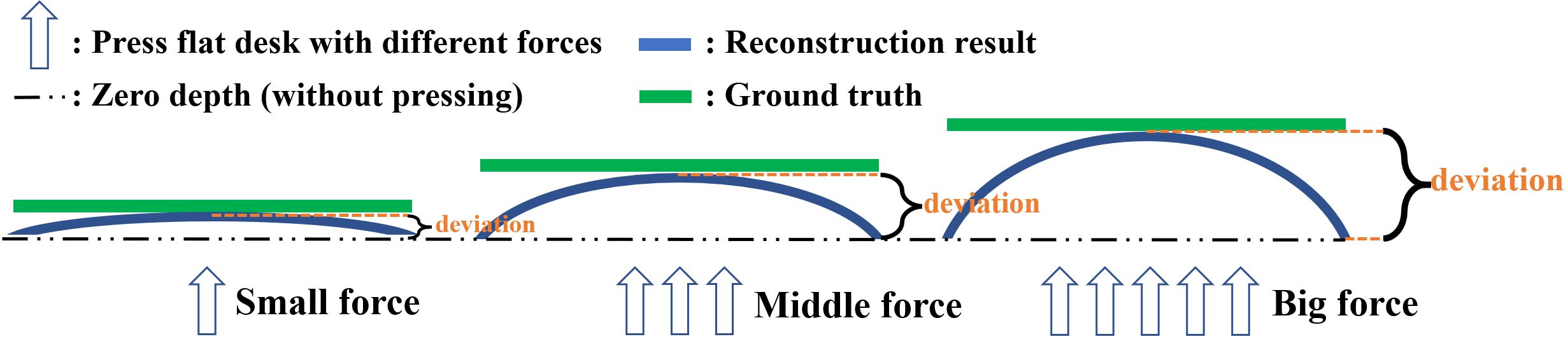}
\caption{\textbf{Different pressures caused different deviations between the center and periphery.}}
\label{fig:pressure-inconsistency}
\end{figure}

According to Wang et al.\cite{wang2021gelsight}, different pressures yield different degrees of gel deformation for the same object, even if the sensor surface is flat. 
Therefore, the shape reflected in the local tactile map does not conform to the real situation, indicating that the accuracy will be affected by accidental errors if the pressure cannot maintain consistency. 
We verify this phenomenon through a more detailed experiment: pressing the sensor on a flat desk using different forces and performing spherical fitting to the reconstructed local tactile maps (all in spherical arc shape) through RANSAC technology. 
As shown in Table \ref{tab:force-and-radius} and Fig. \ref{fig:pressure-inconsistency}, as the pressure increases, the fitted radius decreases, which implies that the depth deviation between the center and periphery becomes more obvious.

Therefore, during global tactile map reconstruction, the errors caused by 1) the arc-shaped sensor surface and 2) the inconsistency of pressure are coupled, further reducing the accuracy. The correction algorithm must consider both factors.

\begin{figure}[b]
\centering
    \includegraphics[width=.45\linewidth]{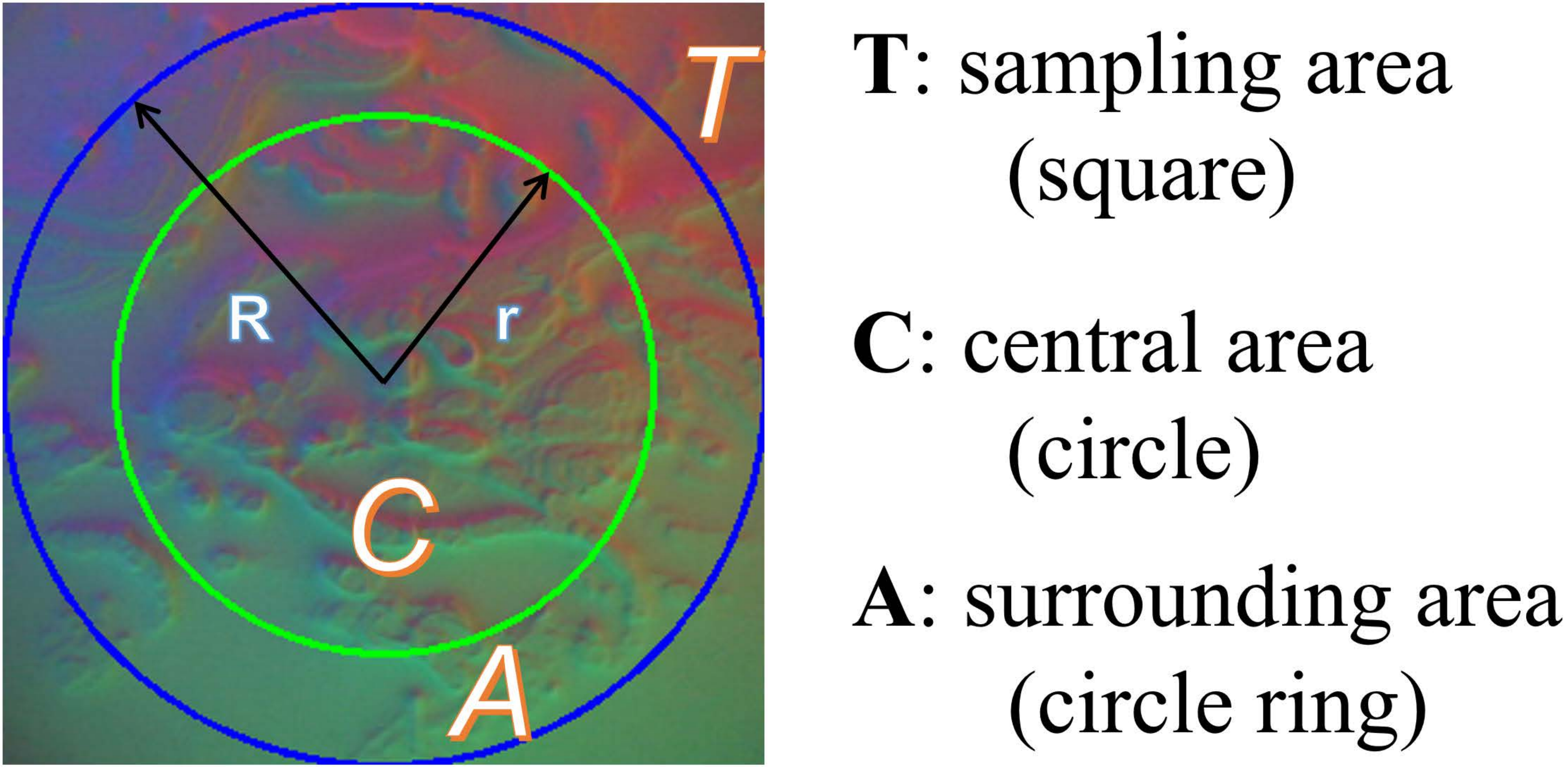}
\caption{\textbf{Force relative magnitude estimation algorithm.} The square area whose side length is $2R$ is the sensor sampling area, denoted as $T$; the central area is a circle with radius $r$, denoted as $C$; the surrounding area is a circle ring between $C$ and a large circle with radius $R$, denoted as $A$. The size of $C$ and $A$ are equal to ensure they are equally important in statistics. In our experiments, $R$ and $r$ are 150 and 110 pixels, respectively.}
\label{fig:forcerelative}
\end{figure}

\subsection{Correction algorithm}
We innovatively propose a correction algorithm that can comprehensively reduce errors in the two aspects mentioned above. 
1) Press the flat desk with a standard force to obtain the standard point cloud $P_{s}$. 
2) Estimate the relative force magnitude coefficient $\alpha$ between each sampling force and standard force.  
3) Generate the correction point cloud $P_{c_i}$ from the standard point cloud $P_{s}$ for each sampled point cloud $P_i$ and then use it to correct the depth of $P_i$. 

Essentially, $P_{s}$ reflects the depth error distribution caused by the arc shape of the sensor surface under specific pressure. 
This distribution is further affected by unstable pressure during sampling. Therefore, it is necessary to estimate the relative magnitude $\alpha_i$ of the sampling pressure between each subsequent sample point cloud $P_i$ and $P_{s}$.
$P_{c_i}$ is generated by adjusting the distribution of $P_{s}$ through $\alpha_i$, such that the correction algorithm can achieve a self-adaptive effect.

\subsubsection{Estimation of the relative magnitude of pressure}
As shown in Fig. \ref{fig:forcerelative}, we divide the sampling area $T$ into two main sub-regions with equal area, $C$ and $A$. We denote the depths of the central and surrounding areas as $d_c$ and $d_a$, respectively, and denote the deviation between them as $\bigtriangleup {d}$. We estimate the relative magnitude coefficient $\alpha$ as follows: 

The depths of areas $C$ and $A$ are estimated by counting the mean depth of the points in their corresponding regions and are denoted as $\hat{d}_c$ and $\hat{d}_a$. The deviation between them is denoted as $\bigtriangleup \hat{d}$. 
This way, the depth deviation of the standard frame and each subsequent sampling frame are denoted as $\bigtriangleup \hat{d}_{s}$ and $\bigtriangleup \hat{d}_i$, respectively, representing the different deviation intensities caused by the inconsistency pressure. 
The relative magnitude coefficient $\hat{\alpha_i}$ between the two frames is estimated as $\hat{\alpha_i} = \frac{\bigtriangleup \hat{d}_i}{\bigtriangleup \hat{d}_{s}}$.

It is worth mentioning that $ \hat{\alpha_i} $ must be estimated for each $P_i$ to improve the depth consistency during the entire sampling process.
When $ \hat{\alpha_i} $ is greater than a threshold (1.1 in our experiments), which implies that the contact surface is possibly curved, we set $ \hat{\alpha_i} $ as the threshold to avoid an inappropriate, excessive correction.
The only assumption is that the correlation between the depth deviation and pressure described in Section \ref{section five-B} is independent of the pressed object surface texture. Thus, the standard point cloud $P_{s}$ collected on the flat desk is suitable for correcting other objects.

\subsubsection{Depth correction}
According to the correlation between the depth deviation and pressure $\bigtriangleup d\propto F$, we perform depth correction for each local tactile map according to Equation \eqref{equ:depthcorrect}
\begin{equation}
\label{equ:depthcorrect}
\begin{aligned}
    \mathbf{P^z_{c_i}} &= \mathbf{P^z_{s}} \times \hat{\alpha_i} \\
    \mathbf{P^z_{i}} &= \mathbf{P^z_{i}} - \mathbf{P^z_{c_i}}
\end{aligned}
\end{equation}
where $\mathbf{P^z}$ is the ${z}$ component (depth distribution) of point cloud data $P$. 
First, we transform the depth distribution of $P_{s}$ by $\hat{\alpha_i}$ to form $P_{c_i}$ with depth deviation $\bigtriangleup \hat{d}_{c_i}$. The correction point cloud $P_{c_i}$ satisfies $\bigtriangleup \hat{d}_{c_i}=\bigtriangleup \hat{d}_{i}$, which implies that it has the same intensity of depth deviation as the local tactile map, $P_i$.
Then, by subtracting the depth values of $P_i$ and $P_{c_i}$, depth correction is completed, making $\bigtriangleup \hat{d}_{i}=0$. 

\section{Global tactile map construction\label{section six}}
This section details how our method fuses multi-frame pressure-corrected local tactile maps into a global tactile map to accurately reconstruct a large-scale object surface. 
We describe the point cloud preprocessing technique and the coarse-to-fine point cloud registration algorithm in Section \ref{subsection:pcp} and then elaborate on the loop-closure detection algorithm and the pose graph optimization algorithm in Section \ref{subsection:tmlcd}.

\subsection{Point cloud preprocessing and registration\label{subsection:pcp}}
Because of the high resolution of the image-based tactile sensor and the fact that only a small area of the sensor surface will be in contact with the object during sampling, the local tactile map is very dense with a low percentage of valid information, making the registration process lengthy and less robust. 

\begin{figure}[t]
\centering
    \includegraphics[width=\linewidth]{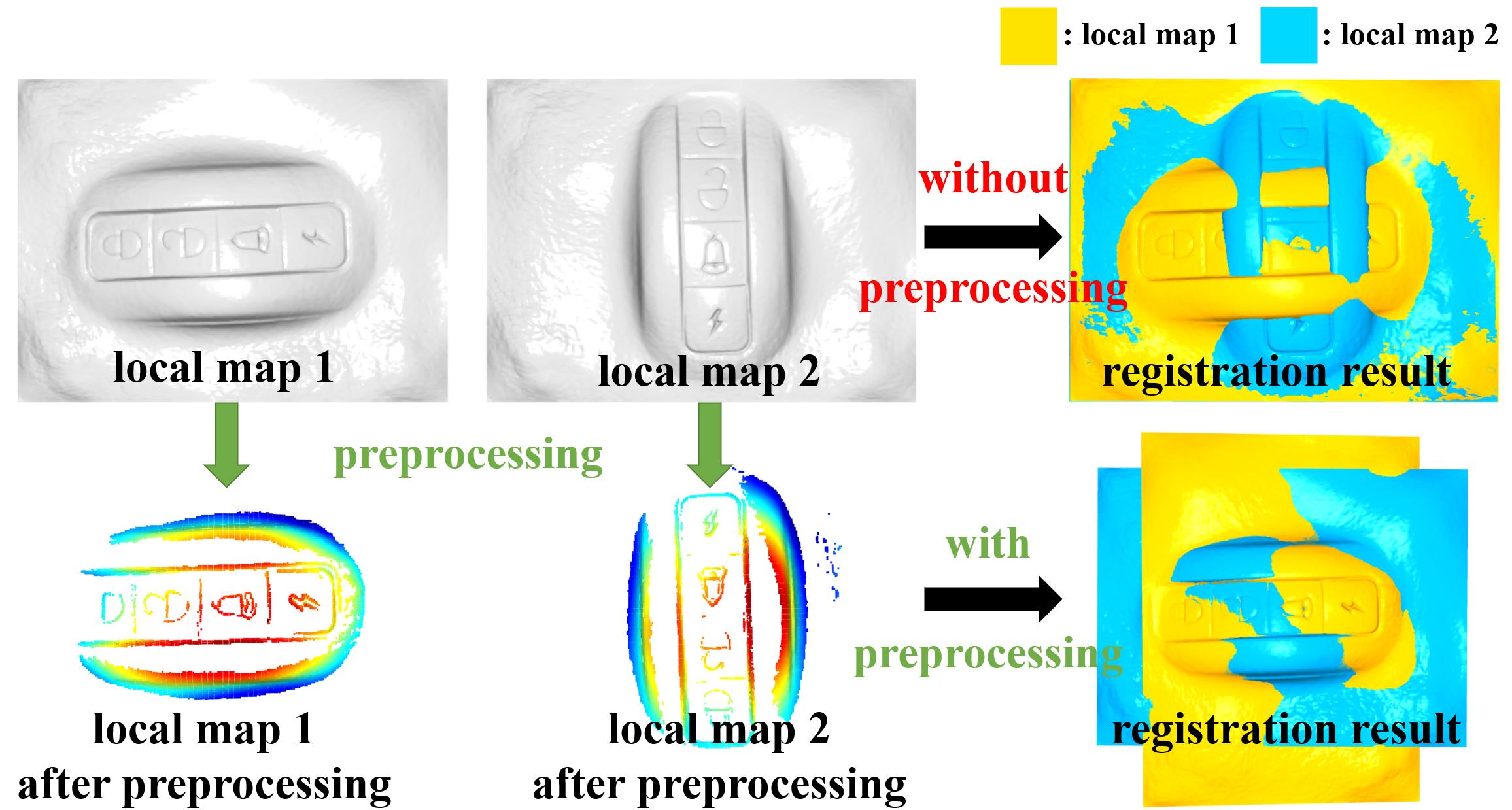}
    \caption{\textbf{Point cloud preprocessing results.} Left and middle: two local tactile maps (top row) and the subset of them whose pitch angle meets the threshold during the ROI extraction process (bottom row); right: the registration results of the two local maps with (bottom row) or without (top row) preprocessing. After preprocessing, the registration result is more accurate.}
    \label{fig:preprocess}
\end{figure}

In our case, before point cloud registration, the preprocessing operations include 1)  voxel downsampling and 2) ROI (region of interest) extraction based on the pitch angle of the surface normal, defined as Equation \eqref{equ:picth_and_yaw}.  
Among them, voxel downsampling helps improve the efficiency of the registration algorithm by properly setting the voxel size (0.3--0.4 mm in our experiments). 
By setting a threshold on the pitch angle (60--80$^{\circ}$ in our experiments), our ROI extraction method detects points with significant deformations and extracts a bounding box for them. The points inside the bounding box are used in the subsequent registration process to increase the robustness of the registration.
As shown in Fig. \ref{fig:preprocess}, the preprocessing module can significantly improve the accuracy of the registration algorithm.

In the registration stage, we implement a coarse-to-fine point cloud registration algorithm based on Open3D, an open-source point cloud library\cite{Zhou2018}. We use the RANSAC algorithm based on FPFH\cite{rusu2009fast} for rough global registration and then use the result as the initial value of the iteration for fine registration based on the point-to-plane ICP algorithm\cite{chen1992object,rusinkiewicz2001efficient}. 

While registering a sequence of local tactile maps, we estimate the transformation matrix between two adjacent frames by registering their point clouds. Then we use the result matrix to align them. 
Repeating this operation allows each local tactile map to be aligned to the same frame. In our case, the first frame is chosen as the base frame, where the position of the tactile sensor is the identity matrix. 

\subsection{Loop-closure detection and pose graph optimization\label{subsection:tmlcd}}
Although the algorithm mentioned in Section \ref{subsection:pcp} can register a sequence of local tactile maps, it is essentially an odometry model with an inevitable accumulative error. When the object's scale is much larger than the effective sampling area of the sensor, the accumulative error significantly reduces the accuracy of the final result. Loop-closure detection algorithms are often used in SLAM to address this problem\cite{bosse2013place,kenshimov2017deep,bai2018sequence}. Algorithms that rely on convolutional neural networks (CNN) for scene re-recognition and loop-closure detection have also been widely used\cite{kenshimov2017deep,bai2018sequence}.

As there is both tactile image and map information during the global tactile map reconstruction, we use the  ResNet152 neural network, pre-trained using the ImageNet dataset\cite{russakovsky2015imagenet}, without its last linear layer as the tactile image descriptor encoder. We determine whether loop-closure occurs based on the cosine similarity of the descriptor vectors defined by Equation\eqref{equ:calc_sim_loop}. $\mathbf{X}$ and $\mathbf{Y}$ are the descriptor vectors of two tactile images with $N$ dimensions.

\begin{equation}
\label{equ:calc_sim_loop}
S_{sim} =  \mathbf{X} \cdot \mathbf{Y} = \sum_{i=0}^{N}\mathbf{X_i}\mathbf{Y_i} 
\end{equation}

\begin{figure}[t]
\centering
    \includegraphics[width=\linewidth]{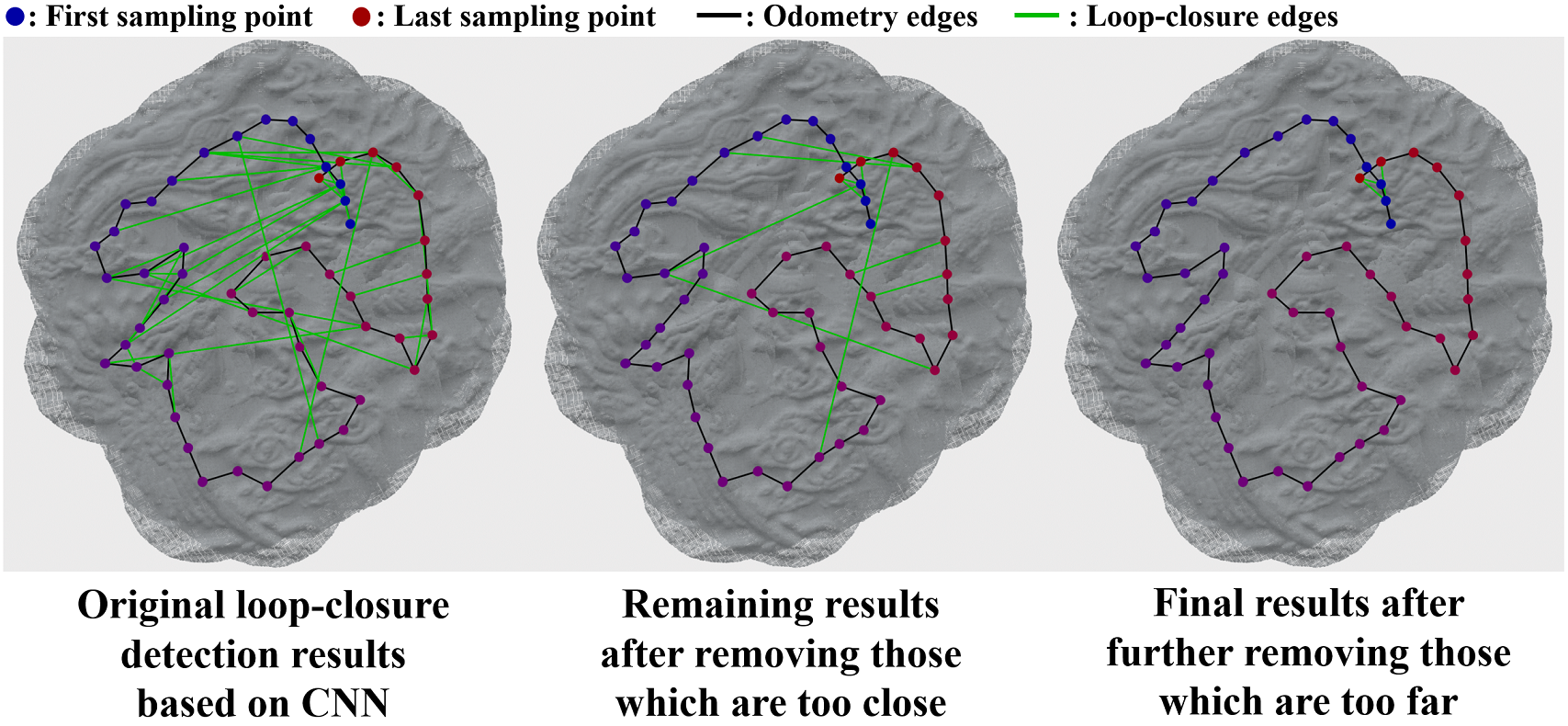}
    \caption{\textbf{Elimination process of false positive loop-closure detection.}
    From the left to right: the original detection results based on CNN (the top 20\% of cosine similarity ranking are retained), results after removing candidate frames that are too close, and final results after further removing the candidate frames that are too far.}
\label{fig:loopclosure}
\end{figure}

Because of the limited size of the sampling area, if the distance between two candidate frames is too large, there will be no common area, which means that they are sampled from two different spots. Therefore, these frames may be false positive detection results. As shown in Fig. \ref{fig:loopclosure}, by filtering out candidate frames whose distance is too close or too far comprehensively considering the efficiency and accuracy, the validity and accuracy of detection are improved, giving the subsequent pose graph optimization more reliable data association information.

The pose graph comprises nodes and edges, where the nodes represent the pose of each frame and the edges represent the relative transformation between two connected nodes. 
In our case, a pose graph is constructed during point cloud registration. 
We initialize the pose graph by setting the pose of the first frame as the identity matrix, coinciding with the world coordinate. 
During the process of point cloud registration, nodes and their adjacent edges in the pose graph are gradually added by the odometry described in Section \ref{subsection:pcp} whereas edges between non-adjacent nodes are added by loop-closure detection. 
Finally, we used the LM (Levenberg--Marquardt) method in the optimization stage to optimize the pose of each node. It is known to have better convergence than the commonly used GN (Gauss--Newton) method.

Overall, loop-closure detection and pose-graph optimization techniques reduce the accumulative error and ensure the accuracy of the surface reconstruction of large-scale objects.

\begin{table*}[!t]
    \centering
    \caption{\label{tab:A_and_r2}The result of fitting line ${y=ax}$ between the ground truth and the predicted value of ${picth}$ and ${yaw}$ angle of each test image. $a$ is the slope of the fitted line and $R^2$ is the coefficient of determination}
    \resizebox{0.65\textwidth}{!}{
    \begin{tabular}{ccccccccccc}
        \hline
        Number & 1 & 2 & 3 & 4 & 5 & 6 & 7 & 8 & 9 & average\\ 
        \hline
        Pitch $a$ & 1.027 & 1.099 & 0.919 & 1.029 & 1.031 & \textbf{1.020} & 1.057 & 1.064 & 1.190 & 1.048\\
        
        Pitch $R^2$ & 0.889 & \textbf{0.907} & 0.799 & 0.870 & 0.897 & 0.793 & 0.793 & 0.810 & 0.862 & 0.847\\ 
        
        Yaw $a$   & 1.045 & 0.953 & 1.020 & 0.963 & \textbf{1.014} & 1.052 & 1.016 & 1.017 & 0.972 & 1.006\\
        
        Yaw $R^2$   & 0.971 & \textbf{0.983} & 0.914 & 0.976 & 0.982 & 0.965 & 0.957 & 0.963 & 0.982 & 0.966\\
        \hline
    \end{tabular}
    }
\end{table*}

\section{Experiments}
In this section, we elaborate on the experimental setup and analyze the result, which comprises four parts. 
First, in Section \ref{subsection:exp_Local}, to illustrate the accuracy of our local tactile map construction algorithm, we describe the training process of the MLP model and quantitatively evaluate it. 
Second, in Section \ref{ransacflag}, we define two indexes:1) the reconstruction flatness of flat objects and 2) the fitted radius of the local tactile map of a hemisphere. 
The first index is used to evaluate the effectiveness of the proposed pressure correction algorithm on flat surfaces. The second index validates the algorithm on a non-flat surface.
Third, in Section \ref{exp:global}, to illustrate the practicability and necessity of the proposed pose graph optimization method, we compare the results of three types of global tactile map construction methods: 1) Tac2S(Odometry), 2) Tac2S, and 3) MC2S (reconstructed by motion capture equipment).  
Finally, in Section \ref{exp-ablation}, we conduct ablation experiments on the core functional modules involved in Tac2Structure and demonstrate the value and necessity of each part.

\subsection{Local tactile map construction\label{subsection:exp_Local}}
\subsubsection{Data collection}
To make the training data as useful and non-repetitive as possible, we collect 60 tactile images with a sphere of diameter 6 mm and one tactile image with nothing pressed to generate 150,000 vectors for training.
Subsequently, a sphere with a diameter of 12 mm is used to collect the test data, and 50,000 vectors are collected.

\subsubsection{Train}
Our multi-layer perceptron model is trained on an Nvidia GTX1060 GPU with 6 GB of memory using the Adam optimizer in Pytorch. We set the learning rate to 0.00112, the batch size to 4000, and the total epoch to 400.

\subsubsection{Evaluation}
We design two performance indexes: (1) the pitch and yaw angles of the surface normal, as in \cite{yuan2017gelsight} and (2) the flatness of the reconstructed point cloud without pressing.

{\bfseries(1) Pitch and yaw angles.}
The surface normal is defined in mathematical form as $\overrightarrow{\mathbf{n}} = (n_{x},n_{y},n_{z})$, where $n_{x}$ and $n_{y}$ are the outputs of the perceptron model and $n_{z}$ is the physical length of one pixel. 
The pitch and yaw angles of the surface normal are defined by Equation \eqref{equ:picth_and_yaw}:

\begin{equation}
\label{equ:picth_and_yaw}
\begin{array}{cc}
&n_{xy}=\sqrt{(n_{x})^2 + (n_{y})^2}\\
&{pitch}=\arctan(\frac{n_{z}}{n_{xy}})\quad,\quad{yaw}=\arctan(\frac{n_{y}}{n_{x}})
\end{array}
\end{equation} 

For all pixels in the nine test tactile images, we perform zero-intercept line fitting on the predicted value and ground truth of $pitch$ and $yaw$, obtaining the slope $a$ and the coefficient of determination $R^2$.
As summarized in Table \ref{tab:A_and_r2}, the mean value of $a$ is approximately 1, and the mean value of $R^2$ is approximately 0.9, which indicates that the perceptron model can accurately estimate the gradient corresponding to each pixel.

{\bfseries(2) Tactile-map flatness without pressing.}
The flatness is defined by the root mean square distance between all points and the theoretical plane $Z=0$, as shown in Equation \eqref{equ:ping}
\begin{equation}
\label{equ:ping}
flatness=\frac{1}{n}\sum_{i=1}^{n}{\Vert \mathbf{p_i}\cdot \overrightarrow{\mathbf{N}}+D \Vert_1}
\end{equation} 
where $\mathbf{p_i}$ is the $i$th point of the local tactile map, $n$ is the number of points in the local tactile map, and $\overrightarrow{\mathbf{N}} = (A,B,C)$. $A,B,C$ and $D$ are parameters in the planar equation ($Ax+By+Cz+D=0$).

The flatness of the reconstruction point cloud of the non-pressing tactile image is calculated to be 0.1869 mm, which implies that our multi-layer perceptron can estimate the depth within millimeter-level accuracy.

\subsection{Tactile map depth correction\label{ransacflag}}
\begin{figure}[b]
\centering
    \includegraphics[width=\linewidth]{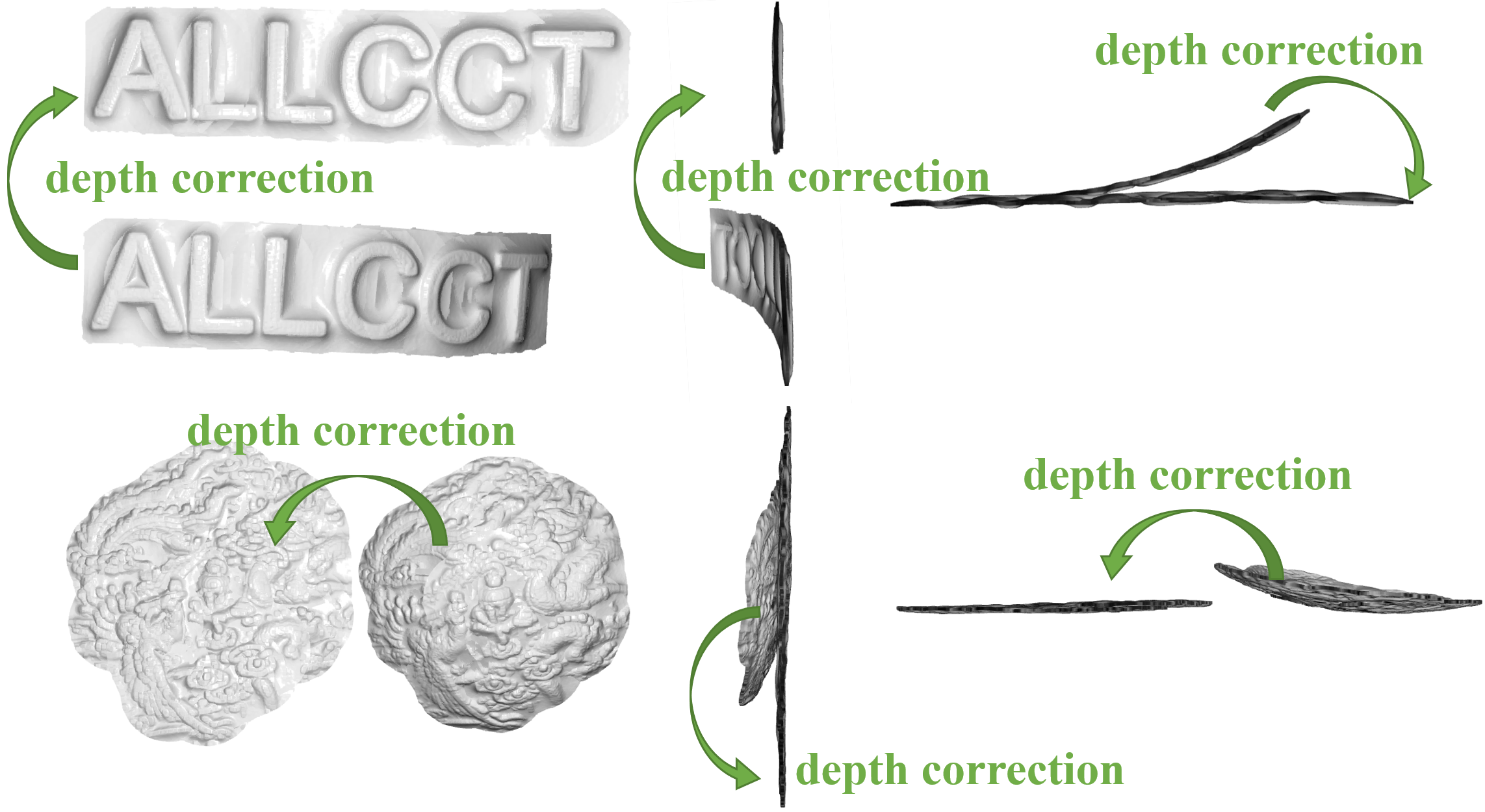}
    \caption{\textbf{Pressure correction point cloud registering results.} From left to right: the front view, side view, and top view of the registration result of the 3D printed `ALLCCT' and `Dragon--Phoenix' with and without depth correction. 
The proposed depth-correction algorithm significantly improves the visual effect of the reconstructed model.}
    \label{fig:correct_and_registering}
\end{figure}

We first evaluate the depth-correction algorithm on flat objects `ALLCCT' and `Dragon--Phoenix.' 
We compare the point cloud registration results (without loop-closure detection and pose graph optimization) of the sequential local tactile maps before and after correction. 
As shown in Fig. \ref{fig:correct_and_registering}, unacceptable bending errors occur in the results before correction, while the results after correction have much better smoothness.

For quantitative comparison, we perform plane fitting on the reconstructed point cloud using RANSAC technology to obtain the parameters in the planar equation. The flatness is the same as that in Equation \eqref{equ:ping}. As shown in Table \ref{tab:ping_compare}, the flatness is significantly improved after correction, indicating that the proposed algorithm can solve the problems caused by the sensor surface appearance and pressure inconsistency.

\begin{table}[!h]
    \centering
    \caption{\label{tab:ping_compare}Flatness before and after correction, Unit: mm}
    \resizebox{.8\linewidth}{!}{
    \begin{tabular}{ccc}
        \hline
        Object & before correction & after correction \\ 
        \hline
        ALLCCT & 2.105 & \textbf{0.807}\\ 
        Dragon--Phoenix & 1.614 & \textbf{0.723} \\
        \hline
    \end{tabular}
    }
\end{table}

To verify the effectiveness on the non-flat surface, we collect 100 tactile images on a 3D printed hemisphere with a known radius (20 mm) using different forces. 
We compare the fitted radius generated from spherical fitting (using the RANSAC technique) on the local tactile maps before and after correction.
As shown in Tabel \ref{tab:radius_compare}, the mean value of the radius after the correction operation is closer to the ground truth, verifying the validity of the algorithm on the curved surface.

\begin{table}[t]
    \centering
    \caption{\label{tab:radius_compare}Fitted radius before and after correction, Unit: mm}
    \resizebox{\linewidth}{!}{
    \begin{tabular}{cccc}
        \hline
        Fitted radius & before correction & after correction & Ground truth\\ 
        \hline
        mean & 17.922 & \textbf{20.569} & \textbf{20}\\ 
        std & 1.300 & 1.695 & ——\\
        \hline
    \end{tabular}
    }
\end{table}

\begin{figure}[b]
    \centering
    \includegraphics[width=\linewidth]{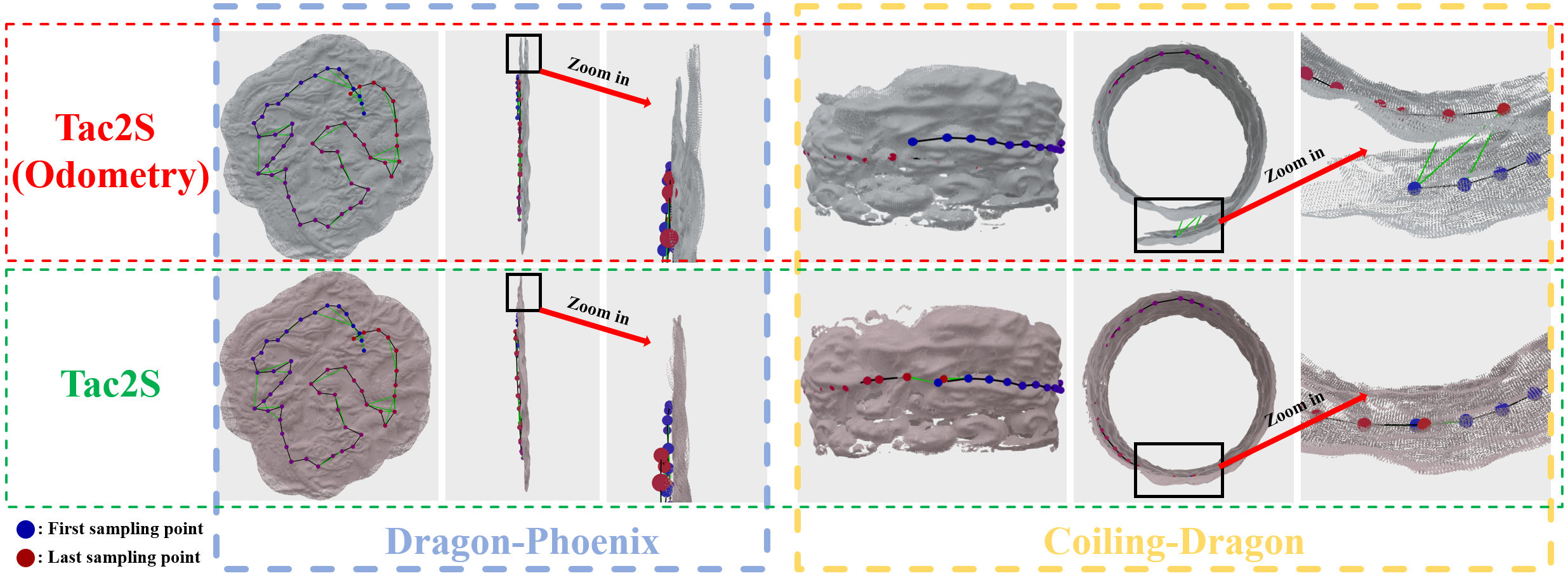}
    \caption{\label{fig:global_map_compare}\textbf{Global tactile map reconstruction results of Tac2S and Tac2S(Odometry).}
    From left to right: the front view, side view, and close-up of the improved area of the global tactile reconstruction results of `Dragon--Phoenix' and `Coiling Dragon.' The pose graph optimization technique significantly decreases the accumulative error of both flat and non-flat objects.}
\end{figure}

\subsection{Global tactile map construction\label{exp:global}}
We compare the global map construction results with and without optimization, named Tac2S and Tac2S(Odometry), respectively, on the `Dragon--Phoenix' and `Coiling Dragon' 3D printed pieces. 
As shown in Fig. \ref{fig:global_map_compare}, the accumulative error of the odometry is very obvious because of the large scale of the objects. Specifically, `Dragon--Phoenix' and `Coiling Dragon' are reconstructed from 55 and 45 local tactile maps, respectively, nearly 20--30 times larger than the sensor sampling area. After the pose graph optimization operation, the drift is significantly reduced, and the reconstruction result is closer to the actual situation of the object surface. 

To quantify the improvement brought by optimization, we design three indexes: 1) the RPE, relative pose error of the first and last frames; 2) flatness (only for flat object `Dragon--Phoenix') and 3) deviation between the reconstructed point cloud and object CAD model. 
To further illustrate the effectiveness of our algorithm, we use a motion capture device (optitrack\cite{optitrack}) to assist in reconstructing the surface texture of the object, as in \cite{li2018end,bauza2019tactile}, named MC2S. 
The second and third indexes evaluate the results of MC2S. 

Specifically, the motion capture system includes four optitrack cameras placed on the four vertices of a square area. 
Five markers are attached evenly to the tactile sensor. 
To stably accept the pose transformation of the tactile sensor during the reconstruction process, we ensure that the makers are always within the common field of view of the four optitrack cameras.

We use the pose obtained by direct point cloud registration between the first and last frames as the true value ($\mathbf{Q}$) and the odometry pose before and after optimization as the estimated value ($\mathbf {P}$) to calculate the RPE, defined by Equation \eqref{equ:rpe}. We only take the length of the translation vector of RPE as the final index, denoted as $ \left \| \mathbf{RPE}_{t} \right \| $.
\begin{equation}
\label{equ:rpe}
\mathbf{RPE} = (\mathbf {Q}_{first}^{-1}\mathbf {Q}_{last})^{-1}(\mathbf {P}_{first}^{-1}\mathbf {P}_{last})
\end{equation} 

The flatness calculation method is consistent with that described in Section \ref{ransacflag}. The deviation between the reconstructed result and CAD model includes the mean value and standard deviation of the distance between them (unit: mm). It is calculated using the open-source software Cloud Compare\cite{cloudcompare}.

The results are listed in Table \ref{tab:pose_graph_v2}. In Tac2S(Odometry), $\left \| \mathbf{RPE}_{t} \right \|$, the flatness and deviation are large. After optimization, all performances improve, indicating the effectiveness of the pose graph optimization algorithm. 
Because the accuracy of MC2S is slightly lower than Tac2S, we believe that the main reason is that optitrack only has millimeter-level accuracy, which is not sufficiently accurate for tactile reconstruction.
Furthermore, the calibration between the optitrack and tactile sensor makes it difficult to achieve millimeter accuracy.

The experimental results verify the necessity and effectiveness of the pose graph optimization technology in our Tac2Structure framework, which can reduce the accumulative error and improve the accuracy of the global reconstruction.

\begin{table}[t]
    \centering
    \begin{threeparttable}[b]
    \caption{\label{tab:pose_graph_v2}Performance comparison table about three global tactile reconstruction methods, Unit: mm}
    \scriptsize
    \begin{tabular}{cccccc}
        \hline
        Object & Group & $ \left \| \mathbf{RPE}_{t} \right \| $ & $flatness$ & $e_{mean}$ & $e_{std}$  \\ 
        \hline
        \multirow{3}{*}{`DP'\tnote{1}}  & Tac2S(Odometry)  & 3.247  & 0.723  & 0.456  & 0.370  \\
                         & Tac2S  & \textbf{0.049}  & \textbf{0.651}  & \textbf{0.416}  & \textbf{0.352}  \\
                         & MC2S  & ——  & 0.717  & 0.440  & 0.381  \\
        \hline
        \multirow{3}{*}{`CD'\tnote{2}}  & Tac2S(Odometry)  & 15.655  & \multirow{3}{*}{——}  & 1.216  & 1.326  \\
                         & Tac2S  & \textbf{0.075}  &   & \textbf{0.390}  & \textbf{0.301}  \\
                         & MC2S  & ——  &   & 0.521  & 0.439 \\
        \hline
    \end{tabular}
    \begin{tablenotes}
     \item[1] `DP' means `Dragon--Phoenix' 3D printed piece
     \item[2] `CD' means `Coiling Dragon' 3D printed piece
    \end{tablenotes}
    \end{threeparttable}
\end{table}

\begin{table}[!b]
\centering
    \caption{\label{tab:Ablation-v2}Table of ablation experiment results, Unit: mm}
\resizebox{.9\linewidth}{!}{
    \begin{tabular}{cccccc}
    \hline
Depth & Pose graph & \multirow{2}{*}{$\left \| \mathbf{RPE}_{t} \right \|$} & \multirow{2}{*}{$flatness$} & \multirow{2}{*}{$e_{mean}$} & \multirow{2}{*}{$e_{std}$} \\
        \multicolumn{1}{l}{correction} & optimization  &      &       &      &     \\ 
        \hline
        \XSolidBrush  & \XSolidBrush   & 7.291          & 1.614           & 5.380   & 5.136 \\
         \CheckmarkBold& \XSolidBrush   & 3.247          & 0.723           & 0.456   & 0.370 \\
         \XSolidBrush  & \CheckmarkBold & 0.056          & 1.749           & 4.847   & 4.613 \\
         \CheckmarkBold& \CheckmarkBold & \textbf{0.049} & \textbf{0.651}  & \textbf{0.416}   & \textbf{0.352} \\
        \hline
    \end{tabular}
    }
\end{table}

\subsection{Ablation experiment\label{exp-ablation}}
Finally, we take the `Dragon--Phoenix' as the experimental subject and conduct an ablation experiment on the core functions of the proposed algorithm (depth correction and pose graph optimization). The results are presented in Table \ref{tab:Ablation-v2}.

The depth-correction algorithm can significantly improve the flatness. In contrast, the pose graph optimization algorithm can reduce the accumulative error of odometry and improve the RPE. 
Combining the two algorithms can reduce the deviation between the reconstruction result and the CAD model. 
This table further shows that the entire algorithm system can reconstruct the object surfaces with millimeter-level accuracy.

\section{Conclusion}
We propose a low-drift framework, Tac2Structure, for the surface reconstruction of large-scale objects, relying only on an image-based tactile sensor. 
Our toolbox helps efficiently complete the annotation of the training dataset. Using the trained MLP model with a Fast Poisson solver, the tactile image can be accurately converted to a local tactile map.
By jointly using the adaptive pressure correction algorithm and point cloud registration algorithm, our method constructs a global tactile map of an object. We use a deep learning-based loop-closure detection algorithm and a pose graph optimization algorithm for large-scale object surface reconstruction to reduce accumulative error and drift. In general, without additional observation equipment, our framework can achieve good accuracy, making the overall operation process simpler, cost cheaper, and applicable to wider application scenarios.

However, similar to the drawbacks of LiDAR odometry in SLAM, our framework fails in scenes without valid textures, making it difficult to reconstruct the surfaces of objects with sharp edges, such as cubes. 
Therefore, our next work is to design and implement a new framework for ``Tactile-IMU" multi-sensor fusion object surface reconstruction. It will make the estimated egomotion of the tactile sensor more robust and accurate in these scenes, greatly expand the applicable scenes and significantly improve the algorithm’s modeling ability.

\bibliographystyle{IEEEtran}
\bibliography{IEEEabrv,ref}

\begin{thebibliography}{10}
\providecommand{\url}[1]{#1}
\csname url@samestyle\endcsname
\providecommand{\newblock}{\relax}
\providecommand{\bibinfo}[2]{#2}
\providecommand{\BIBentrySTDinterwordspacing}{\spaceskip=0pt\relax}
\providecommand{\BIBentryALTinterwordstretchfactor}{4}
\providecommand{\BIBentryALTinterwordspacing}{\spaceskip=\fontdimen2\font plus
\BIBentryALTinterwordstretchfactor\fontdimen3\font minus
  \fontdimen4\font\relax}
\providecommand{\BIBforeignlanguage}[2]{{%
\expandafter\ifx\csname l@#1\endcsname\relax
\typeout{** WARNING: IEEEtran.bst: No hyphenation pattern has been}%
\typeout{** loaded for the language `#1'. Using the pattern for}%
\typeout{** the default language instead.}%
\else
\language=\csname l@#1\endcsname
\fi
#2}}
\providecommand{\BIBdecl}{\relax}
\BIBdecl

\bibitem{newcombe2011kinectfusion}
R.~A. Newcombe, S.~Izadi, O.~Hilliges, D.~Molyneaux, D.~Kim, A.~J. Davison,
  P.~Kohi, J.~Shotton, S.~Hodges, and A.~Fitzgibbon, ``Kinectfusion: Real-time
  dense surface mapping and tracking,'' in \emph{2011 10th IEEE international
  symposium on mixed and augmented reality}.\hskip 1em plus 0.5em minus
  0.4em\relax Ieee, 2011, pp. 127--136.

\bibitem{yuan2017gelsight}
W.~Yuan, S.~Dong, and E.~H. Adelson, ``Gelsight: High-resolution robot tactile
  sensors for estimating geometry and force,'' \emph{Sensors}, vol.~17, no.~12,
  p. 2762, 2017.

\bibitem{donlon2018gelslim}
E.~Donlon, S.~Dong, M.~Liu, J.~Li, E.~Adelson, and A.~Rodriguez, ``Gelslim: A
  high-resolution, compact, robust, and calibrated tactile-sensing finger,'' in
  \emph{2018 IEEE/RSJ International Conference on Intelligent Robots and
  Systems (IROS)}.\hskip 1em plus 0.5em minus 0.4em\relax IEEE, 2018, pp.
  1927--1934.

\bibitem{woodham1979photometric}
R.~J. Woodham, ``Photometric stereo: A reflectance map technique for
  determining surface orientation from image intensity,'' in \emph{Image
  understanding systems and industrial applications I}, vol. 155.\hskip 1em
  plus 0.5em minus 0.4em\relax SPIE, 1979, pp. 136--143.

\bibitem{li2018end}
J.~Li, S.~Dong, and E.~H. Adelson, ``End-to-end pixelwise surface normal
  estimation with convolutional neural networks and shape reconstruction using
  gelsight sensor,'' in \emph{2018 IEEE International Conference on Robotics
  and Biomimetics (ROBIO)}.\hskip 1em plus 0.5em minus 0.4em\relax IEEE, 2018,
  pp. 1292--1297.

\bibitem{bauza2019tactile}
M.~Bauza, O.~Canal, and A.~Rodriguez, ``Tactile mapping and localization from
  high-resolution tactile imprints,'' in \emph{2019 International Conference on
  Robotics and Automation (ICRA)}.\hskip 1em plus 0.5em minus 0.4em\relax IEEE,
  2019, pp. 3811--3817.

\bibitem{wang2021gelsight}
S.~Wang, Y.~She, B.~Romero, and E.~Adelson, ``Gelsight wedge: Measuring
  high-resolution 3d contact geometry with a compact robot finger,'' in
  \emph{2021 IEEE International Conference on Robotics and Automation
  (ICRA)}.\hskip 1em plus 0.5em minus 0.4em\relax IEEE, 2021, pp. 6468--6475.

\bibitem{li2014localization}
R.~Li, R.~Platt, W.~Yuan, A.~ten Pas, N.~Roscup, M.~A. Srinivasan, and
  E.~Adelson, ``Localization and manipulation of small parts using gelsight
  tactile sensing,'' in \emph{2014 IEEE/RSJ International Conference on
  Intelligent Robots and Systems}.\hskip 1em plus 0.5em minus 0.4em\relax IEEE,
  2014, pp. 3988--3993.

\bibitem{wettels2009multi}
N.~Wettels, J.~A. Fishel, Z.~Su, C.~H. Lin, G.~E. Loeb, and L.~SynTouch,
  ``Multi-modal synergistic tactile sensing,'' in \emph{Tactile sensing in
  humanoids—Tactile sensors and beyond workshop, 9th IEEE-RAS international
  conference on humanoid robots}, 2009.

\bibitem{dong2017improved}
S.~Dong, W.~Yuan, and E.~H. Adelson, ``Improved gelsight tactile sensor for
  measuring geometry and slip,'' in \emph{2017 IEEE/RSJ International
  Conference on Intelligent Robots and Systems (IROS)}.\hskip 1em plus 0.5em
  minus 0.4em\relax IEEE, 2017, pp. 137--144.

\bibitem{huh2020dynamically}
T.~M. Huh, H.~Choi, S.~Willcox, S.~Moon, and M.~R. Cutkosky, ``Dynamically
  reconfigurable tactile sensor for robotic manipulation,'' \emph{IEEE Robotics
  and Automation Letters}, vol.~5, no.~2, pp. 2562--2569, 2020.

\bibitem{suresh2021tactile}
S.~Suresh, M.~Bauza, K.-T. Yu, J.~G. Mangelson, A.~Rodriguez, and M.~Kaess,
  ``Tactile slam: Real-time inference of shape and pose from planar pushing,''
  in \emph{2021 IEEE International Conference on Robotics and Automation
  (ICRA)}.\hskip 1em plus 0.5em minus 0.4em\relax IEEE, 2021, pp.
  11\,322--11\,328.

\bibitem{ma2019dense}
D.~Ma, E.~Donlon, S.~Dong, and A.~Rodriguez, ``Dense tactile force estimation
  using gelslim and inverse fem,'' in \emph{2019 International Conference on
  Robotics and Automation (ICRA)}.\hskip 1em plus 0.5em minus 0.4em\relax IEEE,
  2019, pp. 5418--5424.

\bibitem{chaudhury2022using}
A.~N. Chaudhury, T.~Man, W.~Yuan, and C.~G. Atkeson, ``Using collocated vision
  and tactile sensors for visual servoing and localization,'' \emph{IEEE
  Robotics and Automation Letters}, vol.~7, no.~2, pp. 3427--3434, 2022.

\bibitem{dikhale2022visuotactile}
S.~Dikhale, K.~Patel, D.~Dhingra, I.~Naramura, A.~Hayashi, S.~Iba, and
  N.~Jamali, ``Visuotactile 6d pose estimation of an in-hand object using
  vision and tactile sensor data,'' \emph{IEEE Robotics and Automation
  Letters}, vol.~7, no.~2, pp. 2148--2155, 2022.

\bibitem{zhang2014loam}
J.~Zhang and S.~Singh, ``Loam: Lidar odometry and mapping in real-time.'' in
  \emph{Robotics: Science and Systems}, vol.~2, no.~9.\hskip 1em plus 0.5em
  minus 0.4em\relax Berkeley, CA, 2014, pp. 1--9.

\bibitem{mur2015orb}
R.~Mur-Artal, J.~M.~M. Montiel, and J.~D. Tardos, ``Orb-slam: a versatile and
  accurate monocular slam system,'' \emph{IEEE transactions on robotics},
  vol.~31, no.~5, pp. 1147--1163, 2015.

\bibitem{mur2017orb}
R.~Mur-Artal and J.~D. Tard{\'o}s, ``Orb-slam2: An open-source slam system for
  monocular, stereo, and rgb-d cameras,'' \emph{IEEE transactions on robotics},
  vol.~33, no.~5, pp. 1255--1262, 2017.

\bibitem{dai2020rgb}
W.~Dai, Y.~Zhang, P.~Li, Z.~Fang, and S.~Scherer, ``Rgb-d slam in dynamic
  environments using point correlations,'' \emph{IEEE Transactions on Pattern
  Analysis and Machine Intelligence}, vol.~44, no.~1, pp. 373--389, 2020.

\bibitem{DBLP:journals/corr/abs-2111-07723}
\BIBentryALTinterwordspacing
Z.~Wan, Y.~Zhang, B.~He, Z.~Cui, W.~Dai, L.~Zhou, and G.~Huang, ``Enhance
  accuracy: Sensitivity and uncertainty theory in lidar odometry and mapping,''
  \emph{CoRR}, vol. abs/2111.07723, 2021. [Online]. Available:
  \url{https://arxiv.org/abs/2111.07723}
\BIBentrySTDinterwordspacing

\bibitem{doernerpythonfast}
\BIBentryALTinterwordspacing
J.~Doerner. {``Fast poisson reconstruction in python,"}. [Online]. Available:
  \url{https://gist.github.com/jackdoerner/b9b5e62a4c3893c76e4c}
\BIBentrySTDinterwordspacing

\bibitem{romero2020soft}
B.~Romero, F.~Veiga, and E.~Adelson, ``Soft, round, high resolution tactile
  fingertip sensors for dexterous robotic manipulation,'' in \emph{2020 IEEE
  International Conference on Robotics and Automation (ICRA)}.\hskip 1em plus
  0.5em minus 0.4em\relax IEEE, 2020, pp. 4796--4802.

\bibitem{Zhou2018}
Q.-Y. Zhou, J.~Park, and V.~Koltun, ``{Open3D}: {A} modern library for {3D}
  data processing,'' \emph{arXiv:1801.09847}, 2018.

\bibitem{rusu2009fast}
R.~B. Rusu, N.~Blodow, and M.~Beetz, ``Fast point feature histograms (fpfh) for
  3d registration,'' in \emph{2009 IEEE international conference on robotics
  and automation}.\hskip 1em plus 0.5em minus 0.4em\relax IEEE, 2009, pp.
  3212--3217.

\bibitem{chen1992object}
Y.~Chen and G.~Medioni, ``Object modelling by registration of multiple range
  images,'' \emph{Image and vision computing}, vol.~10, no.~3, pp. 145--155,
  1992.

\bibitem{rusinkiewicz2001efficient}
S.~Rusinkiewicz and M.~Levoy, ``Efficient variants of the icp algorithm,'' in
  \emph{Proceedings third international conference on 3-D digital imaging and
  modeling}.\hskip 1em plus 0.5em minus 0.4em\relax IEEE, 2001, pp. 145--152.

\bibitem{bosse2013place}
M.~Bosse and R.~Zlot, ``Place recognition using keypoint voting in large 3d
  lidar datasets,'' in \emph{2013 IEEE International Conference on Robotics and
  Automation}.\hskip 1em plus 0.5em minus 0.4em\relax IEEE, 2013, pp.
  2677--2684.

\bibitem{kenshimov2017deep}
C.~Kenshimov, L.~Bampis, B.~Amirgaliyev, M.~Arslanov, and A.~Gasteratos, ``Deep
  learning features exception for cross-season visual place recognition,''
  \emph{Pattern Recognition Letters}, vol. 100, pp. 124--130, 2017.

\bibitem{bai2018sequence}
D.~Bai, C.~Wang, B.~Zhang, X.~Yi, and X.~Yang, ``Sequence searching with cnn
  features for robust and fast visual place recognition,'' \emph{Computers \&
  Graphics}, vol.~70, pp. 270--280, 2018.

\bibitem{russakovsky2015imagenet}
O.~Russakovsky, J.~Deng, H.~Su, J.~Krause, S.~Satheesh, S.~Ma, Z.~Huang,
  A.~Karpathy, A.~Khosla, M.~Bernstein \emph{et~al.}, ``Imagenet large scale
  visual recognition challenge,'' \emph{International journal of computer
  vision}, vol. 115, no.~3, pp. 211--252, 2015.

\bibitem{optitrack}
\BIBentryALTinterwordspacing
{``OptiTrack - Motion Capture Systems,"}. [Online]. Available:
  \url{https://www.optitrack.com/}
\BIBentrySTDinterwordspacing

\bibitem{cloudcompare}
\BIBentryALTinterwordspacing
{``Cloud Compare: 3D point cloud and mesh processing software,"}. [Online].
  Available: \url{https://www.danielgm.net/cc/}
\BIBentrySTDinterwordspacing

\end{thebibliography}

\end{document}